\documentclass{Interspeech2024}

\interspeechcameraready

\title{CogniVoice: Multimodal and Multilingual Fusion Networks for \\Mild Cognitive Impairment Assessment from Spontaneous Speech}
\name[]{Jiali}{Cheng}
\name[]{Mohamed}{Elgaar}
\name[]{Nidhi}{Vakil}
\name[]{Hadi}{Amiri}

\address{
  University of Massachusetts Lowell, USA
}
\email{\{jiali\_cheng, mohamed\_elgaar, nidhipiyush\_vakil, hadi\_amiri\}@uml.edu}

\usepackage{xcolor}
\usepackage{lmodern}
\usepackage{tcolorbox}
\usepackage{multirow}
\usepackage{xspace}
\usepackage{hyperref}

\newcommand{\method}{CogniVoice\xspace}

\begin{document}

\maketitle

\begin{abstract}

Mild Cognitive Impairment (MCI) is a medical condition characterized by noticeable declines in memory and cognitive abilities, potentially affecting individual's daily activities. 
In this paper, we introduce \method, a novel multilingual and multimodal framework to detect MCI and estimate Mini-Mental State Examination (MMSE) scores by analyzing speech data and its textual transcriptions. The key component of \method is an ensemble multimodal and multilingual
network based on ``Product of Experts'' that mitigates reliance on shortcut solutions. 
Using a comprehensive dataset containing both English and Chinese languages from TAUKADIAL challenge, \method outperforms the best performing baseline model on MCI classification and MMSE regression tasks by 2.8 and 4.1 points in F1 and RMSE respectively, and can effectively reduce the performance gap across different language groups by 0.7 points in F1\footnote{Code: \href{https://github.com/CLU-UML/CogniVoice}{https://github.com/CLU-UML/CogniVoice}.}.

\end{abstract}

\section{Introduction}

Mild Cognitive Impairment (MCI) is a medical condition characterized by noticeable declines in memory, language skills, and logical thinking, and is often observed in the elderly population. MCI is considered as an early indicator or precursor to dementia, but not all cases of MCI progress to this more severe cognitive decline. 
Globally, dementia affects 55 million individuals, ranks as the seventh leading cause of mortality with women being disproportionately impacted, and is a major contributor to disability and dependency among the elderly.\footnote{\href{www.who.int/news-room/fact-sheets/detail/dementia}{www.who.int/news-room/fact-sheets/detail/dementia}.}

Speech 
offers a reflection of cognitive status and has been used as a key digital biomarker for cognitive evaluation. This potential underscores the opportunity for the integration of speech analysis techniques in cognitive health assessment as follows: given speech samples from elderly individuals describing a select set of pictures, the task is to automatically detect the presence of MCI in these individuals and estimate their Mini-Mental State Examination (MMSE) scores through detailed analysis of their speech; the MMSE is a brief 30-point questionnaire test commonly used in clinical settings to assess cognitive function and screen for cognitive loss.

Previous research proposed effective unimodal and multimodal techniques to detect cognitive impairment. 
In~\cite{farzana2023towards}, authors employed feature-based and instance-based domain adaption techniques 
to overcome data sparsity and improve generalizability for dementia detection. 
In~\cite{ustc2023}, several wav2vec models~\cite{baevski2020wav2vec} were fine-tuned on various frequency bands and eGeMAPS~\cite{eyben2015geneva} acoustic features were combined with silence features for Alzheimer's disease recognition. 
In~\cite{santos2017enriching}, transcriptions were transformed into a co-occurrence network with words as nodes, word embeddings as features, and word adjacency in text as edge indicators to better represent short texts produced in MCI assessment. Topological features from the resulting graph (such as page rank and centrality) and linguistics features (such as coherence~\cite{graesser2004coh}) were then integrated to detect MCI.
In \cite{duan_cda_2023}, a contrastive learning approach was developed for detecting Alzheimer’s disease on a small dataset, where negative examples were obtained by randomly removing text segments from transcripts. Other works captured language coherence~\cite{gkoumas2023digital}, explored fusion strategies~\cite{zanwar_what_2023}, mitigated the influence of the examiner using speaker discriminative features~\cite{dawalatabad2022detecting}, used speaker recognition and features from silence segments of speech~\cite{pappagari2020using}, captured linguistic and acoustic characteristics of MCI cases using hand-crafted features derived from domain knowledge~\cite{DBLP:conf/interspeech/BalagopalanERN20}, jointly trained on speech and text data~\cite{chou2023toward}, used paralinguistic features~\cite{chen2023} to detect cognitive impairment.



 




Previous works often focused on mono-lingual models, which demonstrate effectiveness on data from specific languages. Although there has been efforts to develop multilingual systems \cite{tamm2023cross},   
existing methods tend to overfit to spurious correlations or rely on shortcut solutions, 
which undermines models' ability to generalize across languages and patient groups. 



In this work, we develop a novel framework, called \method, to extract multimodal and multilingual features from speech inputs and their corresponding transcripts to predict MCI and cognitive test outcomes in elderly English and Chinese speakers. The key contribution of the paper is a systematic approach to ensemble multimodal and multilingual networks based on ``product of experts'' (Section~\ref{sec:poe}). The approach effectively encourages learning robust multimodal and multilingual speech and text features, reduces overfitting, and mitigates reliance on shortcut solutions in the above tasks. 



\method outperforms the best performing baseline on MCI classification and MMSE regression tasks by 2.8 and 4.1 points in F1 and RMSE respectively. Existing methods have significant performance disparity between patient groups of different languages. \method can effectively reduce such performance gap by 0.7 point in F1.\looseness-1




\begin{figure*}[t]
  \centering
  \includegraphics[width=0.95\linewidth]{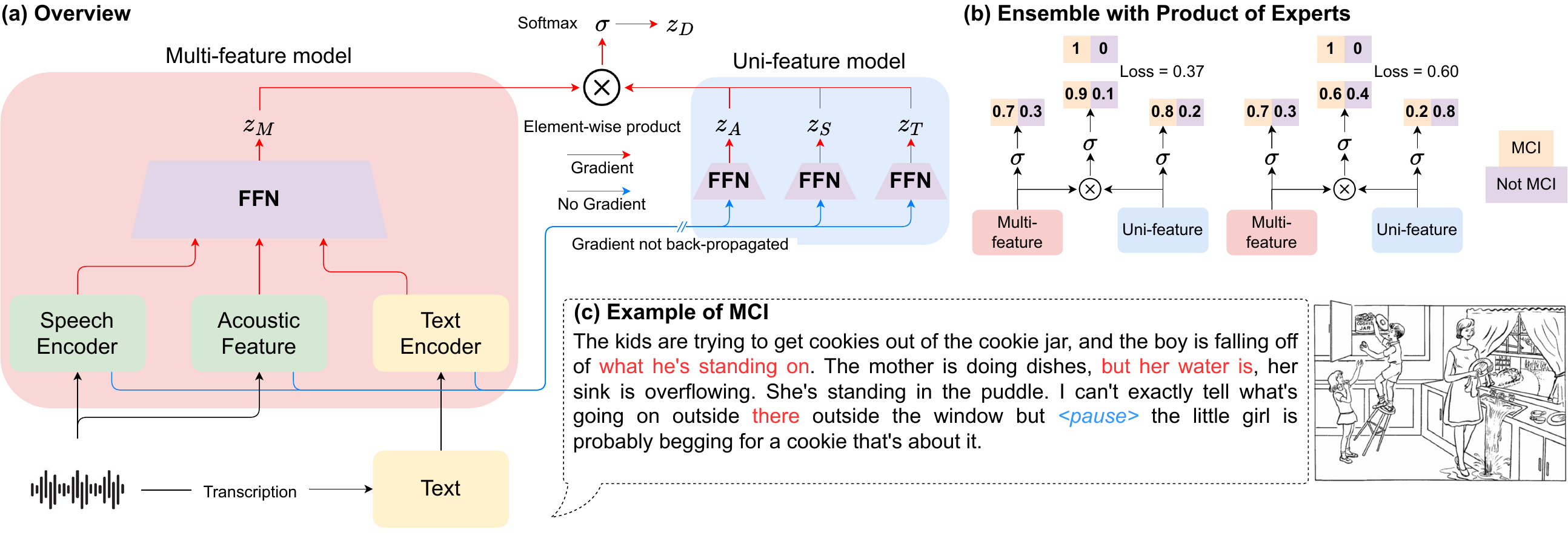}
  \caption{Schematic diagram of the proposed approach. 
  (a) Overall architecture of \method: a multi-feature model is fused with different uni-feature models through product-of-experts (PoE) to mitigate overfitting and prioritize true task signals over spurious features. (b) mitigating overfitting with PoE. (c) Example of speech provided by a patient diagnosed with MCI describing the cookie theft picture~\cite{borod1980normative} for MCI assessment. The description lacks coherence and clear transitions between events, lack to remember or identify the objects and use vague language highlighted in the text, which are important signs of cognitive impairment. Non-MCI patients often narrate a sequence of events with clarity, identify various elements in the picture, provide detailed description, recall objects and events already described.
  }
  \label{fig:model}
\end{figure*}

\section{\method}\label{sec:approach}

\subsection{Overview}
Given $D = \{x^0, x^1,\dots,x^{N-1}\}$, a dataset of speech samples of elderly individuals describing a select set of pictures, we aim to train a model $f$ that correctly predicts the presence of MCI in these individuals and estimates their MMSE scores.

To diagnose MCI through speech, clinicians pay close attention to several key signs including word retrieval and repetition issues, change in language use, difficulties with attention and focus, confusion about time and place, and mood swings. 
These indicators can often be extracted from speech data, using acoustic and textual features. 
However, due to the limited training data, standard training can easily lead to overfitting, perhaps through learning spurious correlations between superficial features and class labels. 
This prioritize some features, e.g. some of acoustic features, while ignoring the other indicators that clinicians consider, and can lead to significant performance degradation on unseen test samples. To mitigate the overfitting issue and encourage model to focus on genuine and robust features, we propose to train the model $f$ using Product-of-Experts (PoE)~\cite{clark-etal-2019-dont,karimi-mahabadi-etal-2020-end}. 

Our approach consists of several components depicted in Figure~\ref{fig:model}. Given an input speech sample, we extract features using transformers for speech and its corresponding text as well as acoustic features obtained from DisVoice~\cite{disvoice1,disvoice2,disvoice3,disvoice4,disvoice5,disvoice6,disvoice7}. A standard training approach concatenates all features and optimizes the cross-entropy loss, see the {\em multi-feature model} in Figure~\ref{fig:model}. To model potential shortcut signals within each feature sets, we propose to train with PoE, applied to the multi-feature model and several {\em uni-feature models}, which predict the labels using only one set of features separately. Our approach obtain ensemble logits using the multi-feature and uni-feature models, see element-wsie product in Figure~\ref{fig:model}(a). In addition,  Figure~\ref{fig:model}(b) shows how PoE can reduce the loss for samples correctly predicted using both multimodal and unimodal inputs, and increase the loss for samples that cannot be accurately predicted using one of the modalities, which allows for identifying and mitigating weaknesses in the model's predictive capabilities. Therefore, the resulting ensemble logits can account for the spurious correlations in the dataset, while also being regularized to mitigate overfitting.

\subsection{Mitigating Spurious Feature with PoE}\label{sec:poe}
Given a speech $S$ and its transcribed text $T$, we first feed them into the multi-feature model to yield an initial prediction as follows:
\begin{equation}
    z_M = f_{\mathrm{Multi}}(S, T).
\end{equation} 

For clarity, we illustrate the PoE concept using a single uni-feature model as an example:
we can model the spurious features in $T$ by feeding the extracted text features into a separate feed-forward network (FFN) to predict class labels:
\begin{equation}
    z_T = \mathtt{FFN}_T \Big(f_T(T) \Big),
\end{equation}
which is then combined with the multi-feature model's prediction $z_M$ using element-wise product (in log space for efficiency and stability) to encourage $f_{\mathrm{Multi}}$ to prioritize robust features over spurious ones:
\begin{equation}
    log z_F = log z_M + log z_T.
\end{equation}

Compared to $z_M$, the resulting prediction $z_F$ contains guidance from $f_{\mathrm{Uni}}$, which is used to adjust the standard cross-entropy loss for training a more robust MCI classifier. To avoid confusing the multi-feature model $f_{\mathrm{Multi}}$, the gradient from the uni-feature model is not back-propagated through $f_{\mathrm{Multi}}$, which is a standard practice in previous work on PoE~\cite{karimi-mahabadi-etal-2020-end,rubi}. During inference, we only use $f_{\mathrm{Multi}}$ to make predictions.

\noindent \textbf{Why PoE works} We present two examples in Figure~\ref{fig:model}(b). When the uni-feature model is confident about a correct prediction, PoE can increase the confidence of the multi-feature model, resulting in a smaller loss. This indicates that the multi-feature model has learned the sample and is therefore less adjusted. However, when the uni-feature model is confident about a wrong prediction, the prediction from the multi-feature model has a more balanced confidence toward possible classes resulting in a larger loss. In this case, the input sample will have a larger contribution on the update of model parameters. We also provide two interpretations of PoE:
(a): PoE regularizes model's prediction when model relies on spurious correlations, rather than learning causal relations; this regularization prevents model from overfitting; and
(b): PoE dynamically adjusts the weight of training samples, which can be viewed as a dynamic curriculum that adaptively re-weights the contribution of each training example at every training iteration. 

\subsection{Multilingual and Multimodal Features}\label{sec:features}
We extract the following multi-lingual features of different modality from speech sample $S$ and its transcribed text $T$ to facilitate MCI prediction.

\noindent \textbf{Transformer-based speech features}: 
speech samples contain rich signals and indicators for MCI. We employ a Whisper encoder~\cite{radford2023robust} $f_S$ to extract features from each speech sample $S$, denoted as $f_S(S)$. 

\noindent \textbf{Transformer-based text features}: other rich features related to MCI may exist in the transcripts. These include lexical repetitiveness and topical coherence, which may not be directly captured by the speech processing model. To capture such textual features, we transcribe the speech $S$ into text $T$ and employ a language-specific BERT~\cite{devlin-etal-2019-bert} encoder $f_T$ to extract features from $T$, denoted as $f_T(T)$.

\noindent \textbf{Acoustic features}: we use the following acoustic features in our model: {\em static} features for the entire utterance and {\em dynamic} features obtained frame-by-frame, from DisVoice. \textit{Phonation features} related to tone, pitch, loudness, and quality to capture jitter and shimmer.
\textit{Phonological features} compute the  posteriors probabilities of phonological classes from audio files for several groups of phonemes considering the mode and manner of articulation. 
\textit{Articulation features} compute features related to sounds and their production from continuous speech.
\textit{Prosody features} compute features from continuous speech focusing on duration, energy and fundamental frequency.    
\textit{Representation learning features} are computed using convolution and recurrent auto-encoders. Features are based on training the model to find MSE loss between decoded and input spectrogram, and extracting features from the last layer of the encoder. 
Each type of features is represented by a vector. We compute the average for each feature, resulting in a fixed size vector $f_A(S)$,\footnote{We use a size of 10 in our experiments.} where $f_A$ is the DisVoice feature extractor.

\noindent \textbf{Feature fusion}: given all the available features $f_S(S)$, $f_T(T)$, and $f_A(S)$, we feed them into a FFN to combine them and compute interactions between different types of features, extracting useful signals for downstream MCI prediction. Overall, the multi-feature model computes 
\begin{equation}
    z_M = f_{\mathrm{Multi}}(S, T) = \mathtt{FFN}_M \Big([f_S(S); f_T(T); f_A(S)] \Big),
\end{equation} where $z_M \in \mathbb{R}^2$ denotes the logit and $[;]$ denotes vector concatenation.

\noindent \textbf{Final model}: we extend the PoE to all three types of features, including $z_S$, $z_T$ and $z_A$. The resulting PoE is be obtained as follows:
\begin{equation}
    log z_F = log z_M + log z_S + log z_T + log z_A,
\end{equation} after which $z_F$ is used to compute the cross-entropy loss.



\begin{table}[t]
\scriptsize
\centering
\caption{Statistics of the TAUKADIAL dataset. Count indicates number of speech files in each category.}
\label{tab:data_stats}
\begin{tabular}{l c c c c c c}
        \toprule
         & \multicolumn{2}{c}{\textbf{Gender}}    & \textbf{Avg.} & \multicolumn{2}{c}{\textbf{Classification}} & \textbf{Avg.} \\
         \textbf{Lang.} & \textbf{F} & \textbf{M} & \textbf{Age}  & \textbf{MCI} & \textbf{NC}                  & \textbf{MMSE} \\
         \midrule
         \textbf{En} & 123 & 63  & 71.2 & 123 &  63 & 28.3 \\
         \textbf{Zh} & 114 & 87  & 74.2 & 99  & 102 & 26.3\\
         \midrule
         \textbf{Total}   & 237 & 150 & 72.7 & 222 &  165 & 27.2 \\
         \bottomrule
\end{tabular}
\end{table}

\begin{table*}[htb]
\small
\centering
\caption{Results of MCI classification in F1 score and unweighted average recall (UAR) averaged across k-fold cross validation. Columns M and F show the performance only on male and female subjects. Columns EN and CN show the performance depending on the spoken language. Best performance is in \textbf{Bold} and second best is \underline{underlined}.}
\label{tab:mci_res}
\vspace{-5pt}
\begin{tabular}{l|rrrrr|rrrrr}
& \multicolumn{5}{c|}{F1 ($\uparrow)$} & \multicolumn{5}{c}{UAR ($\uparrow)$} \\
& Avg. & M &  F &  En & Zh & Avg. & M & F & En & Zh \\
\toprule
 Whisper        & 81.3 & \underline{82.0} & 79.9 & \underline{80.6} & 72.2 & 73.5 & \underline{80.8} & 69.5 & 57.9 & \textbf{79.0} \\
 AST            & 71.0 & 77.3 & 66.4 & 76.0 & 53.2 & 62.7 & 79.2 & 55.0 & 56.7 & 63.5 \\
 XLSR-53        & 62.9 & 72.2 & 56.9 & 74.2 & 42.9 & 53.7 & 63.8 & 48.2 & 55.0 & 52.9 \\
 XLS-R (0.3B)   & 75.5 & 78.0 & 72.1 & 79.5 & 61.1 & 60.5 & 67.9 & 55.5 & 56.7 & 57.2 \\
 \midrule
\method w/o PoE & \underline{81.7} & 80.9 & \underline{82.1} & \underline{80.6} & \underline{79.5} & \underline{73.6} & 78.5 & \underline{73.2} & \textbf{58.9} & \underline{78.7}  \\
\method         & \textbf{84.1} & \textbf{87.8} & \textbf{82.3} & \textbf{81.3} & \textbf{81.7} & \textbf{75.1} & \textbf{84.1} & \textbf{72.3} & \underline{58.4} & 77.5  \\
\bottomrule
\end{tabular}
\end{table*}

\begin{table*}[t]
\small
\caption{Results of MMSE prediction averaged across k-fold cross validation.} 
\label{tab:reg_res}
\vspace{-5pt}
\centering
\begin{tabular}{l|rrrrr|rrrrr}
& \multicolumn{5}{c|}{RMSE ($\downarrow)$} & \multicolumn{5}{c}{R\textsuperscript{2} ($\uparrow)$} \\
& Avg. & M &  F &  En & Zh & Avg. & M & F & En & Zh \\
\toprule
Whisper      & 9.30 & 9.44 & 9.13 & 9.79 & 8.73 & -8.33 & -27.59 & -6.98 & -104.29 & -4.10 \\
AST          & \underline{3.42} & \underline{2.55} & \underline{3.95} & \underline{3.95} & \underline{3.95} & \underline{-0.3} & -0.8 & -0.4 & -0.4 & -0.4 \\
XLSR-53      & 3.72 & 2.76 & 4.34 & 4.34 & 4.34 & -0.3 & \underline{-0.7} & \underline{-0.3} & \underline{-0.3} & \underline{-0.3} \\
XLS-R (0.3B) & 5.21 & 4.31 & 5.81 & 5.81 & 5.81 & -2.3 & -4.6 & -2.2 & -2.2 & -2.2 \\
\midrule
\method w/o PoE & \textbf{2.34} & \textbf{2.03} & \textbf{2.39} & \textbf{1.21} & \textbf{2.93} & \textbf{0.45} & \textbf{0.02} & \textbf{0.50} & \textbf{-0.06} & \textbf{0.43} \\
\bottomrule
\end{tabular}
\end{table*}


\section{Experimental Setup}

\subsection{Dataset}

We use the MCI dataset from TAUKADIAL Challenge 2024~\cite{taukadial}. The dataset contains speech data from 129 participants, 62 (48.1\%) are English speaker and 67 (51.9\%) Chinese speaker. Each participant is asked to describe three pictures leading to 387 data points. Age and gender of the participants are also provided. Each description of an image is considered as a datapoint and is labeled to determine the presence of MCI in an individual in conjunction with mini mental status examination (MMSE) score, which indicates the severity of the MCI. Table \ref{tab:data_stats} shows the statistics of the dataset.

\subsection{Baseline and settings}
We compare our method to the following baselines:
\begin{itemize}
    \item \textbf{Whisper}~\cite{radford2023robust}: An encoder-decoder transformer model trained on speech transcription and translation using spectrogram input. We fine-tune the encoder for classification and regression.
    \item \textbf{Wav2Vec 2.0}~\cite{baevski2020wav2vec}: A speech representation extraction model, trained using a self-supervised masked-token-prediction objective.
    \item \textbf{Audio Spectrogram Transformer} (AST)~\cite{gong21b_interspeech}: Speech encoder using patched spectrogram input. 
    \item \textbf{XLSR-53}~\cite{conneau2021unsupervised}: A Wav2Vec 2.0 that learns a shared space for quantized speech units of 53 languages.
    \item \textbf{XLS-R}~\cite{Babu2021XLSRSC}: A Wav2Vec 2.0 model trained with 436k hours of speech in 128 languages. 
\end{itemize}

We train all methods for 10 epochs with a learning rate of 1e-5 and L2 regularizer $\lambda$=0.01 on an A100 GPU.

\subsection{Evaluation}
We use F1 and Unweighted Average Recall (UAR) score to evaluate the classification performance. For the regression task, we use Rooted Mean Squared Error (RMSE) and R\textsuperscript{2}. We also report the performance of different subgroups, including male/female and English/Chinese patients. Due to the small size of the dataset, we adopt a stratified $k$-fold ($k$=10) cross validation to compute average validation score over $k$ folds for comparison.


\begin{figure}[t]
  \centering
  \includegraphics[width=0.95\linewidth]{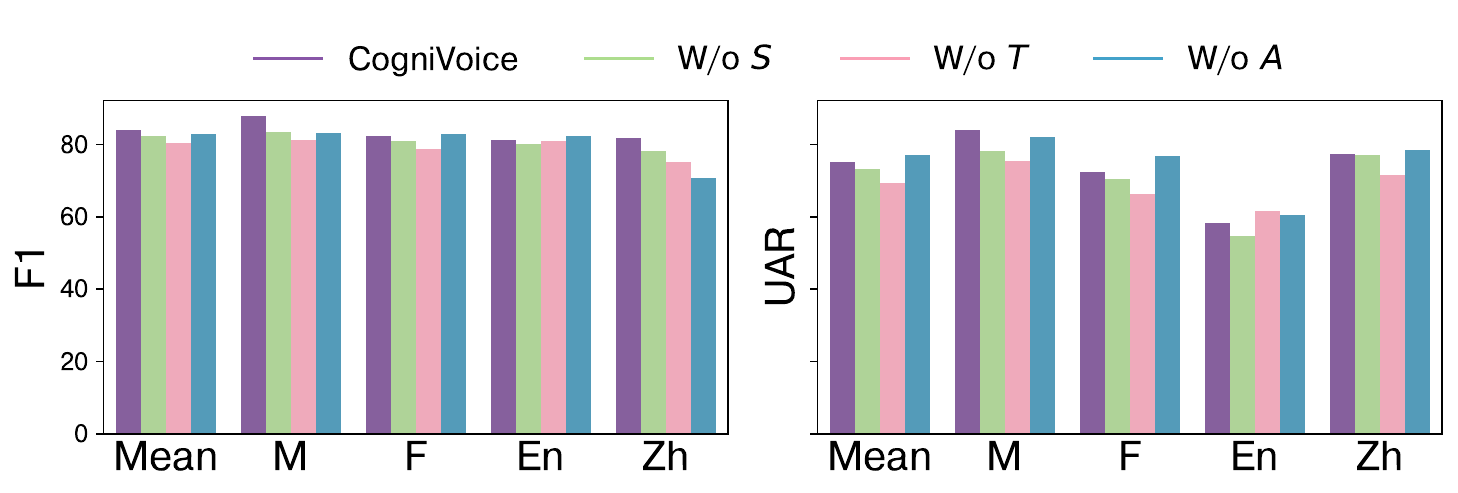}
  \caption{Ablation study on multimodal features.}
  \label{fig:ablation}
\end{figure}

\subsection{Main Results}
On MCI prediction task, \method achieves an F1 score of 84.1, outperforming Whisper-Tiny, AST,  XLSR-53, and XLS-R  by 2.8, 13.1, 21.2,  and 8.6 absolute points, respectively. On MMSE regression task, \method achieves an RMSE of 2.34, outperforming Whisper-Tiny, AST, XLSR-53, and XLS-R by 6.96, 1.08, 1.38, and 2.87 absolute points, respectively. 


\noindent \textbf{\method reduces prediction disparity across patient groups}: 
As shown in the Table \ref{tab:mci_res}, overall, compared to other models \method performs better across all groups. All the models have higher F1 score on English compared to Chinese language. Moreover, all the models performs well for male speakers compared to female speakers. Comparing Whisper-tiny against XLS-R(0.3B) model, increasing the size of the model does not lead to better MCI classification. Similarly, as shown in the Table \ref{tab:reg_res} \method performs better compared to all other models across all groups on the regression task, where we observe similar patterns in terms of RMSE.  


\noindent \textbf{Effect of PoE}: results in Table~\ref{tab:mci_res} show that PoE can  increase the overall F1 score from 81.7 to 84.1 (+2.4), and UAR from 73.6 to 75.1 (+1.5). Meanwhile, PoE can also increase the worst case F1 score in subgroups. With PoE, the worst case F1 are 82.3 and 81.7 for gender and language subgroups, respectively, higher than 80.9 and 79.5 when PoE is not incorporated. Nevertheless, the worse case UAR degrades when PoE is incorporated. In addition, PoE can reduce the performance disparity across different languages, where the gaps drop from 1.1 to 0.4 and 19.8 to 19.1 for F1 score and UAR, respectively. Across gender subgroups, however, PoE may cause higher performance gap than Non PoE.

\noindent \textbf{All features contribute}: Figure~\ref{fig:ablation} show that removing speech features from the model (\textit{W/o S}) results in degradation of F1 score 
by 1.7 
points. 
When text feature are removed (\textit{W/o T}), F1 score drops from 84.1 to 80.4. Interestingly, the F1 score on English-speaking patients only drops by 0.3 point, significantly smaller than Chinese-speaking patients, which has a drop of 6.5 points. On the other hand, the English-speaking patients has an increased UAR by 3.2 points, while the overall UAR and all other groups exhibit degradation. 
Removing DisVoice features (\textit{W/o A}) decreases the F1 score by 4.6 points on male patients while increases it by 0.6 point on female patients. The F1 score increases by 1.2 points on English-speaking patients, while drops by a large margin of 10.9 points on Chinese-speaking patients, indicating that the critical contribution of language-specific text encoder.
These results highlight the contribution of the collected multimodal and multilingual features.

\section{Conclusion}
We proposed a novel model to extract multimodal and multilingual features from speech and transcribed text to predict MCI and MMSE regression score. Our model uses an ensemble approach based on Product Of Experts to effectively learn robust speech and text features and show reduced prediction disparity across patient groups.

\newpage



\bibliographystyle{IEEEtran}
\bibliography{reference,anthology_part2}

\end{document}